\documentclass[11pt,a4paper]{article}
\usepackage[hyperref]{naaclhlt2018}
\usepackage{times}
\usepackage{bbm}
\usepackage{latexsym}
\usepackage{graphicx}
\usepackage{enumitem}
\usepackage{url}

\aclfinalcopy 

\author{Jingyun Liu \\
	McGill University \\
	{\tt \small jingyun.liu@mail.mcgill.ca} \\\And
	Jackie C.K.Cheung \\
	McGill University \\
	{\tt \small jcheung@cs.mcgill.ca} \\\And
	Annie Louis \\
University of Edinburgh\\
{\tt \footnotesize alouis@inf.ed.ac.uk}}

\newcommand\thoughts[1]{}     

\title{What comes next? Extractive summarization by next-sentence prediction}

\date{}

\begin{document}
\maketitle

\begin{abstract}
Existing approaches to automatic summarization assume that a length limit for the 
summary is given, and view content selection as an optimization problem to 
maximize informativeness and minimize redundancy within this budget. This 
framework ignores the fact that human-written summaries have rich internal 
structure which can be exploited to train a summarization system. 
We present \textsc{NextSum}, a novel approach to 
summarization based on a model that predicts the next sentence to include  
in the summary using not only the source article, but also 
the summary produced so far. We show that such a model 
successfully captures summary-specific discourse moves, and leads
to better content selection performance, in addition to automatically
predicting how long the target summary should be. 
We perform experiments on the New York Times 
Annotated Corpus of summaries, where \textsc{NextSum} outperforms lead and content-model 
summarization baselines by significant margins. We also show that the lengths of 
summaries produced by our system correlates with the lengths of the 
human-written gold standards.	
\end{abstract}

\section{Introduction}

Writing a summary is a different task compared to producing a longer
article. As a consequence, it is likely that the topic and 
discourse \emph{moves} made in summaries differ from those in regular 
articles. In this work, we present a powerful extractive summarization system 
which exploits rich summary-internal structure to 
perform content selection, redundancy reduction, and even predict
the target summary length, all in one joint model. 


Text summarization has been addressed by 
numerous techniques in the community \cite{nenkova-mckeown:2011}.
For extractive summarization, which is the focus of this paper, a popular task setup is
to generate summaries that respect a fixed length limit. In the 
summarization shared tasks of the past Document 
Understanding Conferences (DUC\footnote{\url{http://duc.nist.gov/}}), these limits
are defined 
in terms of words or bytes. As a result, 
much work has framed summarization
as a constrained optimization problem, in 
order to select a subset of sentences with desirable summary qualities 
such as informativeness, coherence, and 
non-redundancy within the length budget
\cite{gillick-favre:2009,lin-bilmes:2011,kulesza-taskar:2011}. 

One problem with this setup is that it
does not match many real-world summarization settings. 
For example, writers can tailor the length of 
their summaries to the amount of noteworthy content 
in the source article. Summaries created by news editors for archives, 
such as the New York Times Annotated Corpus \cite{NYTcorpus}, exhibit a variety of lengths. 
There is also evidence that in the context of web search, people prefer summaries 
of different lengths for the documents in search results depending on the type 
of the search query \cite{kaisser-hearst-lowe:2008}. More generally, current systems
focus heavily on properties of the source document to learn to identify
important sentences, and score the coherence of sentence transitions.
They reason about the content of summaries primarily for 
purposes of avoiding redundancy, and respecting the length budget. 
But they ignore the idea that it might actually be useful
to learn content structure and discourse planning for summaries from large 
collections of multi-sentence summaries. 

This work proposes an extractive summarization system that focuses on 
capturing rich summary-internal structure. Our key idea is that 
since summaries in a domain often follow some 
predictable structure, a partial summary or set of summary sentences should help 
predict other summary sentences. 
We formalize this intuition in a model 
called \textsc{NextSum}, which selects the next summary sentence based not only on 
properties of the source text, but also on the previously 
selected sentences in the summary. An example choice is shown in Table~\ref{tab:example}.
This setup allows our model to capture \emph{summary-specific discourse and topic transitions}. 
For example, it can learn to expand on a topic that is already 
mentioned in the summary, or to introduce a new topic. It can 
learn to follow a script or discourse relations that are 
expected for that domain's summaries. It can even learn to predict the end of the summary, avoiding 
the need to explicitly define a length cutoff.


The core of our system is a next-sentence prediction component, which is a 
feed-forward neural network driven by features capturing the prevalence of 
domain subtopics in the source and the summary, sentence importance in the source, and 
coverage of the source document by the summary so far. 
A full summary can then be generated by repeatedly predicting the next sentence until 
the model predicts that the summary should end.

Since summary-specific moves may depend on the domain, we first explore domain-specific summarization on 
event-oriented news topics (War Crimes, Assassinations, Bombs) from the
New York Times Annotated Corpus \cite{NYTcorpus}. We also train a domain-general 
model across multiple types of events.
\textsc{NextSum} predicts the next summary sentence with remarkably high accuracies, 
reaching 67\% compared to a chance accuracy of 9\%. 
The generated summaries outperform the lead baseline 
as well as domain-specific summarization baselines without requiring explicit 
redundancy check or a length constraint. Moreover, 
the system produces summaries of variable lengths which correlate with how long human summaries are for the same texts.

\begin{table}
\centering{
\begin{footnotesize}
\begin{tabular}{|@{~~}p{7cm}@{~}|} \hline
{\bf Summary so far}\\   
{[S]} After a sordid campaign overshadowed by the killing last month of a leading
liberal politician, the citizens of St. Petersburgh, Russia's second largest
city, voted in record numbers Sunday, and today's preliminary results indicated
that they had improved the fortunes of the city's embattled democratic alliance. \\ \hline
{\bf Correct next summary sentence}\\
{[A]} The biggest winner on Sunday was Yabloko, a liberal party led by a presidential
aspirant, Grigory A. Yablinsky, whose candidates in 24 districts scored well enough
to move to the final round. \\ \hline
{\bf Incorrect as next sentence}\\
{[B]} Yabloko, which has long considered St. Petersburg as its stronghold, was even opposed
by a party that called itself Yabloko-St Petersburg.\\ \hline
\end{tabular}
\end{footnotesize}
\caption{Example of a partial summary {[S]}, and two sentences from the same
source article. Both {[A]} and {[B]} are about the same entity, but {[A]} is 
clearly a logical next sentence to continue {[S]} when compared to {[B]}.}
\label{tab:example}
}
\end{table}

\section{Related work}


Many  approaches to extractive summarization are unsupervised, and 
focus on the role of word frequency and source document representation
for selecting informative and non-redundant content \cite{Drago04,Rada04,nenkova06}.
More recently, supervised approaches are popular, which view content 
selection as a sentence-level binary classification problem, typically 
using a neural network \cite{cheng-lapata:2016,Nallapati2017}.

\vspace{1mm}
{\bf Using source structure.} 
Source structure is a common cue for summarization. Relative word frequency and
position of sentences are standardly used in many systems.
Discourse- and graph-based summarization techniques explicitly focus on computing
document structure
\cite{marcu98,louis-joshi-nenkova:2010:SIGDIAL,christensen-etal:2013}. 
Other techniques include learning probabilistic topic models over 
source articles within a domain to capture subtopics and transitions between them
 \cite{barzilay04,haghighi2009,P13-1039}. 
But, the use of structure from summaries is less explored. 

\vspace{1mm}
{\bf Using summary structure.} 
Actually, almost all systems maintain some representation of the partial summary at a timestep. 
At the very least, it is needed for respecting a length limit and for preventing redundancy.
Even in recent neural network based extractive summarization, a representation of the
summary so far has been proposed to allow redundancy checks \cite{Nallapati2017}. 
However, current methods do not focus on capturing rich summary discourse and 
content structure. 


Recent abstractive neural summarization models based on encoder-decoder 
frameworks actually have greater scope for capturing summary structure 
and content. The use of techniques such as attention and pointer mechanisms 
can be viewed as a form of summary structure 
modelling \cite{rush-chopra-weston:2015,Nallapati2016abstractive,see-liu-manning:2017,Socher2017}.
However, because such systems currently operate at the word level, 
these mechanisms are mostly used for 
handling issues such as grammaticality, out-of-vocabulary items, 
predicate-argument structure, and local coherence. 
By contrast, we aim to capture higher-level transitions in the contents of a summary.


\vspace{1mm}
{\bf Next-sentence prediction.} 
The way we learn summary structure is by training a 
module for next summary sentence 
prediction. A parallel idea can be found in the form of  next-utterance prediction in 
retrieval-based dialogue systems \cite{jafarpour2010,wang-EtAl:2013:EMNLP,lowe-etal:2016}. 
There have also been recent attempts at predicting the
next sentence in text. The skip-thought model \cite{Kiros2015} is trained to 
predict a sentence from its neighbouring
sentences to produce sentence representations. 
\newcite{CLSTM2016} and \newcite{pichotta:acl16} evaluate
neural language models on next-sentence and event prediction. 
In contrast, we aim to predict the next output sentence
within the tangible application of summarization.



\section{\textsc{NextSum} model overview}

We first present the key ideas,  and 
the next section explains how we implement the model. 

\textsc{NextSum} comprises two components, a next-sentence 
prediction system, and a summary
generation module. The first is a supervised system trained to 
select the next summary sentence, given a set of candidate sentences
from the source, and the summary generated so far. \textsc{NextSum}'s 
generation component builds a summary by making repeated calls to the
next-sentence predictor.

\subsection{Predicting the next summary sentence}
The next-sentence predictor is trained on a corpus of source articles
and their gold-standard summaries written by humans. In this work, we focus on 
single-document summarization. 

Consider a source article $X = \{s_1,..,s_M\}$ containing $M$ sentences, and a gold-standard
\emph{extractive} summary $G=g_1..g_N$, a \emph{sequence} 
of $N$ sentences. Since $G$ is extractive, $G \subseteq X$.  

In \textsc{NextSum}, summaries are created by adding one sentence
at a time. Let $Y_T=y_1..y_T$ be the partial summary at timestep $T$; $Y_T$ has $T$ sentences.  
At time $T+1$, the goal of \textsc{NextSum} is to score a set of candidate sentences 
from the source, $C_{T+1} =\{s_1,..,s_K\}$, $C_{T+1} \subseteq X$ and find the 
best next sentence to follow $Y_T$. Let the gold-standard next sentence be $g_{T+1}$.
The set $C_{T+1}$ may either be all of the source sentences
which have not yet been included in the summary, or be limited to a smaller 
size $K \ll M$.
For now, assume that all the unselected source sentences are in the candidate set,
and thus $g_{T+1} \in C_{T+1}$.

The model selects the next summary
sentence from $C_{T+1}$ such that:

\vspace{-2mm}
\[ {\hat{y}}_{T+1} = {\arg\max}_{s_i \in C_{T+1}} Pr(s_i|X,Y_T;\theta )\]
\vspace{-4mm}

When there is a tie, the earlier sentence in the article is selected.
In this work, $Pr(s_i|X,Y_T;\theta )$ is estimated
by a neural network parameterized by $\theta$. 
Recall that the oracle next sentence $g_{T+1}$ is in 
$C_{T+1}$. Hence one approach 
to learn the parameters of $Pr(s_i|X,Y_T;\theta )$ is to frame it as a
binary classification problem where the label 
for sentence $g_{T+1} \in C_{T+1}$ is 1, and 0 for all $s_w \in C_{T+1}$ where $s_w \ne g_{T+1}$.
We implement this classifier using a
feed-forward neural network which takes the encoded
representations of ($X$, $Y_T$ and $s_i$), and
outputs the probability of label 1, $p_{s_i}$, which we use as
$Pr(s_i|X,Y_T;\theta)$. The loss for the classification at timestep $T+1$ is the binary cross-entropy loss:

\vspace{-2mm}
\[ L = -\log p_{g_{T+1}} -\sum_{s_i \in C_{T+1}; s_i \ne g_{T+1}} \log(1-p_{s_i})\].
\vspace{-3mm}

\newcommand{\eos}{\langle\textrm{EOS}\rangle}

One of the special features of \textsc{NextSum} is that we
model the end of the summary within the same setup. To do so, we introduce a
special sentence $\eos$ (End of Summary) to mark the end of every gold-standard
summary, i.e. $G=y_1..y_N\eos$. 
In the model, $\eos$  is included in candidate sets at every timestep. This inclusion allows 
the model to learn to discriminate between selecting a 
sentence from the source
versus ending the summary by picking the $\eos$ marker.  Thus our candidate
set is in fact $C'_{T+1} = C_{T+1} \cup \{\eos\}$.




\subsection{Summary generation}
After the next sentence prediction model is trained, it can
be used to generate a complete summary for a source article.
The model performs this task
by iteratively predicting the next sentence until $\eos$ is selected. 
Note that, unlike previous work, the generation component is not given the target length of the summary. 




\section{Implementing \textsc{NextSum}}
In this section, we explain how we select the candidate set, what features we use in the
neural network for next sentence prediction, and the design of the generation component.

\subsection{Candidate selection}
\label{sec:candidateselection}

Some source articles are very long, which 
means that $C'_{T+1}$ can contain many candidate sentences if we take all of the unselected sentences as candidates. In practice, we limit the size of $C'_{T+1}$ in order to reduce the search
space of the model, which improves running time. 

In the single-document scenario, the source text sentences are in 
a natural discourse, and thus in a logical and temporal order. Hence, it is 
not unreasonable to assume that a good summary is a subsequence of the source. 
Given this assumption, suppose the last sentence chosen for the summary is 
$s_j$ at timestep $T$, then we consider the $K$ sentences
in the source immediately following $s_j$ as the candidate set at time $T+1$. 

During development, 
we found that when $K=10$, the gold-standard next summary sentence is in the 
candidate set 90\% of the time, and is present 80\% of the time when using $K$=5. 
Based on this empirical support for the subsequence hypothesis, 
we use $K=10$ plus the end of summary marker for all the experiments in this paper, for a total candidate set 
size of 11. For comparison, a source 
article in our corpus has on average 33 sentences, and the maximum 
is as high as 500 sentences. 
During training, when fewer than 10 sentences remain, we randomly sample 
other sentences from the entire article to ensure having 
enough negative samples. The model is trained on balanced dataset 
by downsampling, and tested on the distribution 
where each candidate set has size 11.


\subsection{Features for next sentence prediction}

We have a source document $X=\{s_1..s_M\}$ with  $M$ sentences, $Y_T=y_1..y_T$ is a partial summary at time $T$, 
 and let $s$ be a sentence (or $\langle$EOS$\rangle$) in the
candidate set ${C'}_{T+1}$. \textsc{NextSum}'s next sentence prediction relies on 
computing $Pr(s|X,Y_T;\theta)$ using a feedforward neural network with parameters $\theta$. 
This network learns from rich feature-based
representations of $X$, $Y_T$, $s$, and their interactions.

\vspace{1mm}
{\bf Domain subtopics.} These features are based on topics induced from a 
large collection of documents in the same domain as the source article. 

These topics are obtained
using the content-model approach of \newcite{barzilay04}. 
The content model is a Hidden Markov 
Model (HMM), where the states correspond to topics, and transitions between them indicate how 
likely it is for one topic to follow another. The emission distribution from a state is a bigram language model indicating what 
lexical content is likely under that topic. Each sentence in the article is emitted by one state (i.e., one topic).
The probability of an article $T=s_1...s_N$ under a HMM with $M$ 
states $\{\textrm{topic}_1,..,\textrm{topic}_M\}$ is given by:

\vspace{-1ex}
\[\sum_{\textrm{topic}_1..\textrm{topic}_n} \prod_{i=1}^{N} P(\textrm{topic}_i|\textrm{topic}_{i-1})P(s_i|\textrm{topic}_i)\]

Content models can be trained in an unsupervised fashion to maximize the log likelihood of 
the articles from the domain. We choose the number of topics on a development set.\footnote{The 
number of topics are between 10 and 27 for the domains in our corpus.} 



Once trained, the model can compute the most likely state sequence for sentences in the source document, and in the partial summary, using Viterbi decoding. 
Based on the predicted topics, we compute a variety of features:
\vspace{-1ex}
\begin{itemize}[noitemsep]
  \item the proportion of source sentences assigned to each topic
	\item the proportion of sentences in the partial summary assigned to each topic
  \item the most likely topic of the candidate $s$ given by ${\arg\max}_{i \in \textrm{Topics}} P(\textrm{topic}_i|s)$
	\item the emission probability of $s$ from each topic
  \item the transition probability between the topic of the previous summary sentence $y_T$, and the topic of $s$,
$P(\textrm{topic}(s)|\textrm{topic}(y_T))$  

  \item a global estimation of observing the candidate $s$, $P(s)= \sum_{i \in \textrm{Topics}}P(s|\textrm{topic}_i)$
\end{itemize}



\vspace{1mm}
{\bf Content.} We compute an encoding of source, summary so far, and the candidate sentence by averaging the pretrained word2vec
embeddings \cite{mikolov:naacl2013} (trained on Google News Corpus) 
of each word in the span  (900 features in total, 300 each for the source, summary so far, and the candidate). We also add features
for the 1,000 most frequent words in the training articles, in order to encode their presence in $s$, and in the 
sentence previous to $s$ in the source article, i.e. ($s-1$). Similarly, for $s$ and $s-1$, we record 
the presence of each part-of-speech tag and named entity.
We expect these features for $s$ and $s-1$ are useful for 
predicting $\eos$, since the last sentence in a summary might contain some lexical cues.

\vspace{1mm}
{\bf Redundancy.} These features calculate the degree to which the candidate sentence
overlaps with the summary so far. They include
$\textit{sim}(s, y_t)$ for $t = T$, $T-1$, $T-2$ (3 features), where
$\textit{sim}(p,q)$ is computed using cosine similarity between count vector 
representations of the words in $s$ and $y_t$. We also include 
the number of overlapping nouns and 
verbs between $s$ and $Y_T$ (2 features).


\vspace{1mm}
{\bf Position.} The position of a sentence in the source document is an important indicator
for content selection and is widely used in systems. We 
indicate the position in the source of the last generated summary sentence $y_T$ 
(as one of 5 bins, the size of each bin depends on the length 
of the source article). We also indicate the 
position of the candidate sentence, and 
its distance to 
$y_T$ in the source (normalized by the length of the source).

\vspace{1mm}
{\bf Length.} We include features for the length of the
source, both as number of sentences, and number of words (binned into 5 bins). We also include 
the number of sentences and words in the summary so far. The length measures for the partial summary are not binned.

\vspace{1mm}
{\bf Coverage.} These features compute how much of the source will be covered by 
the summary when a candidate sentence is added to it. 
We use the KL divergence between the source and candidate 
summary when $s$ is included in it:
$D_{KL} (X\|Y_T \cup\{s\})$ where the distribution of $X$ and $Y_T$ are
 unigram language models.

\vspace{1mm}
{\bf Sentence importance.} We also indicate the individual importance of a candidate 
sentence. The frequency of a word in the source is known to be a strong feature
for importance \cite{msrrep05}. With this intuition, we include the 
$ \frac{1}{|s|}\sum_{w \in s} \textrm{uni}_X(w)$ where $w$ is a token in the 
candidate sentence, and $\textrm{uni}_X(w)$ is the
unigram probability of $w$ in the source $X$. 

We also use a separate pre-trained model of word importance. This model feeds the context of a 
target word (the two words before and two words after) into a LSTM model which 
outputs the probability of the target word appearing in a summary. The 
importance score of a sentence is then the average and maximum of the predicted scores of each word in the sentence. This model is trained on the same training and development data sets. 

\subsection{Summary generation}
To generate the full summary, the model employs a greedy method that simply calls the next-sentence prediction module 
repeatedly until $\eos$ is selected.  We also tried beam search decoding for a more globally optimal 
sequence of sentences, but we found in preliminary experiments that this search did not improve our  results.

\section{Data}

We hypothesize that next-sentence prediction is more likely to be successful in event-oriented domains
(describing events as opposed to explanations and opinions). 
Moreover, summary-specific moves may be more prominent and learnable from 
summary-article pairs within specific domains compared to a general corpus. 

So we create three domain-specific and one domain-general dataset, all focusing on events.
We use the New York Times Annotated Corpus (NYtimes) \cite{NYTcorpus} since it provides topic metadata, 
has thousands of article-summary pairs on different topics, and summaries are not written to set lengths. 
We selected three topics: 
``War Crimes and Criminals'' ({\sc crime}), ``Assassinations and Attempted Assassinations'' ({\sc assassin.}), and
``Bombs and Explosives'' ({\sc bombs}). We also create a 
more general dataset ({\sc mixed}) by randomly sampling from all the three domains. 

We sample a similar number of articles across each domain, and  randomly
split each domain into 80\% training, 10\% development and 10\% test data. Table~\ref{tab:datastats}
shows the sizes of these datasets. 

We use the Stanford CoreNLP toolkit \cite{CoreNLP} to tokenize, 
segment sentences, and assign part of speech tags to all the texts.

\subsection{Length of articles and summaries}

As previously mentioned, summaries are often written to express the summary-worthy 
content of an article, and not restricted to an arbitrary length. This property can be
seen in our data (Table~\ref{tab:lengthdetails}). 

The NYTimes summaries are abstractive in nature and range from
a minimum of 2 words\footnote{Sometimes just the caption to a photo, not very common.} 
to as many as 278 words. The last column of the table gives the Kendall Tau  
correlation (corrected for ties) between the length of the source and the summary. There is a
significant positive correlation, implying that the length of the
 article is indicative of its information content. This finding motivates us to include
the length of the source article as a feature for next sentence prediction, though we note
that the source length by itself is not enough to determine the summary length without 
doing further analysis of the source content. 


\begin{table}
\centering{
\small{
\begin{tabular}{|l| r r r|}\hline
{\bf Domain} & {\bf Train.} & {\bf Dev.} & {\bf Test}\\\hline
{\sc crime}   &   986         &     123       & 123 \\
{\sc assassin.} & 1,087         &     136       & 136 \\
{\sc bombs}   & 1,440         &     180       & 180 \\
{\sc mixed}   & 1,600         &     200       & 200 \\\hline
\end{tabular}
\caption{Number of article-summary pairs in our data.}
\label{tab:datastats}
}}
\end{table}

\begin{table}
\small{
\begin{tabular}{|l| p{0.3cm} p{0.5cm} p{0.3cm} | p{0.3cm} p{0.5cm} p{0.3cm}| r|}
\hline
{\bf Domain}          & \multicolumn{3}{|c|}{\bf Source} & \multicolumn{3}{|c|}{\bf Summary} & {\bf Tau}\\
                      & {min} & { max} & {avg} & { min} & { max} & {avg} & \\ \hline
{\sc crime}           &   4  &  8,300  & 648   & 2   &  236  &  51 & 0.548 \\
{\sc assassin.}       &   3  &  6,081  & 705   & 3   &  226  &  60 & 0.481 \\
{\sc bombs}           &  48  &  7,808  & 874   & 15  &  278  &  82 & 0.343 \\
{\sc mixed}           &   3  &  7,819  & 815   & 3   &  278  &  81 & 0.358 \\ \hline
\end{tabular}
\caption{Min, max and average lengths (in words) of source articles and abstracts. Tau is the Kendall Tau
correlation between the length of source and abstract.}
\label{tab:lengthdetails}}
\end{table}


\subsection{Obtaining extractive summaries}
\label{sec:oracle-extracts}

The summaries from NYTimes are abstractive in nature. Our system is extractive, and for training the next sentence
selection from the source, we need a mapping between the abstractive summary and the sentences in the 
source article. Note that we create these extractive 
summaries only for training our model. We will evaluate \textsc{NextSum}'s 
output by comparing with the abstractive human summaries as is standard practice. 

We map each sentence in the
abstract to the most similar sentence in the source article. Let $A=a_1..a_n$ be the sequence of sentences in the abstract. 
For each $a_i$, we find $y_i = \arg\max_{s_j \in X}\cos(a_i,s_j)$ where $X$ is the set of source sentences, and 
$\cos(p,q)$ is the cosine similarity between the word unigrams of $p$ and $q$. 

The sequence $Y=y_1..y_n$ corresponding to 
 $A=a_1..a_n$ forms the gold standard extractive summary. Since the extractive summary mirrors
the sequence of content in the abstract, the structure of the summary is preserved, allowing our
next sentence prediction system to be trained on the extractive sequence of sentences. It is also for this reason that we do not
use summarization datasets such as the CNN/Daily Mail corpus \cite{hermann:nips2015} where summaries 
are three-sentence \emph{highlights}, and do not have any discernible discourse structure as a whole.  

\section{Experiments}
We first evaluate our model intrinsically on the 
next-sentence prediction task, then test its performance on
the full summary generation problem.


\subsection{Next-sentence prediction}

Here, the goal is to select the best sentence to follow the 
partial summary from a candidate set of 11 options (see Section~\ref{sec:candidateselection}).
For evaluating this part of our system, we assume that we have 
oracle partial summaries; i.e., the partial summary at timestep $T$, $Y_T=y_1..y_T$
is the same as the gold summary sequence up to time $T$. The question is how 
well we can predict the 
next sentence in this sequence from the candidate set $C'_{T+1}$. The correct answer is the 
sentence in the gold standard at position $T+1$. 
The prediction at each timestep is a separate classification example. 

Recall that we framed the machine learning problem as one of binary classification. We thus
present two sets of results: a) on the binary task, and 
b) on the final choice of one sentence from the candidate set (among the 11 candidates). 
In task (a.), the binary evaluation, the model discriminates among the 2 classes by thresholding 
at $p_s>0.5$. 
The best setting has 4 hidden layers, each layer comprising between 500 to 1,500 neurons. We trained the model by backpropagation using the Adam optimizer \cite{Kingma:2014} for up to 75 epochs.  Hyperparameters were tuned on the development set.
The choice of a
final sentence, Task (b.), is made by picking the candidate sentence with highest $p_s$.

Table~\ref{tab:nextpredresults} shows the accuracy on binary 
classification task and 1-of-11 task, on the different domains. 
In the 1-of-11 task, the expected chance-level accuracy is roughly 9.1\%, since we force every candidate set to have size 11. Our next-utterance
prediction system's accuracy is between 60 to 67\% on the different domains, showing that there
are distinctive clues on summary internal structure and content, which can be learned by a model. 
Note also that the accuracy numbers are consistent across all domains and the mixed case
indicating that the patterns are fairly domain-general within event-oriented documents. 

These evaluations are somewhat idealistic in that the model has access to oracle 
partial summaries during prediction. We next evaluate \textsc{NextSum} on the full summarization task.

\subsection{Summary generation}

We developed two versions of our system. 
Previous methods of summary content selection assume a 
fixed length limit. To compare against these systems, 
in one version of our model, \textsc{NextSum}$_{L}$, 
the length limit is provided as a constraint.
If, after the model generates a summary sentence, 
the word count exceeds the given length, we stop generation 
and truncate the last sentence so the summary is within the length limit. 
The second version, \textsc{NextSum}, is the full model which predicts the 
summary length. Both systems have no access to the oracle partial summary, and use their
 own previous decisions to construct the partial summary. 

We evaluate all the summaries by comparing them with gold-standard \emph{abstracts}
using ROUGE \cite{lin2004}.\footnote{The settings are ROUGE-1.5.5.pl 
-n 2 -x -m -2 4 -u -c 95 -r 1000 -f A -p 0.5 -t 0 -d.} We use ROUGE-2 F-score, as \textsc{NextSum} 
generates summaries of varied length. 

\begin{table}
\centering{
\small{
	\begin{tabular}{|l | c | c|}
          \hline
		{\bf Domain}    & {\bf Binary}& {\bf 1 of 11}\\ 
                                & {\bf Accuracy  (\%)}   & {\bf Accuracy (\%)}\\ \hline
		{\sc Crime}     &  74.5      & 67.2\\ 
		{\sc Assassin.} &  71.7      & 63.8\\ 
		{\sc Bombs}     &  73.5      & 60.0\\ 
                {\sc Mixed}     & 73.1      & 60.9\\ \hline
                Random          &  50.0      & 9.0\\\hline
	\end{tabular}
	\caption{Results on next sentence prediction task.}
        \label{tab:nextpredresults}
}}
\end{table}

\subsubsection{Baselines and comparison systems}
In all these systems, the target length of the summary is given as a constraint. 
We set the length $k$ to the average length (in words) of summaries in the training dataset 
for each domain 
(Table \ref{tab:lengthdetails}). 

{\bf Lead} takes the first $k$ words from the source article. 
For single-document extractive summarization, the lead is a very strong 
baseline which many systems fail to beat \cite{DUC2002overview}. 

{\bf CHMM} is the approach used by \newcite{barzilay04} 
for extractive summarization using content models. 
CHMM computes an importance score for each topic $v$. This score is a probability computed by: 
1) counting the articles in the training set where $v$ appears in both the article and its summary,
2) and normalizing by the number of articles containing $v$. 
To generate a summary, the model ranks the topics in order of decreasing importance, and adds
one sentence from the source for each topic (breaks ties 
randomly if multiple sentences decoded into the same topic). The generation stops upon reaching the length limit. 
This method scores the summary-worthy nature of sentences based solely on their topic. 

{\bf Transition} is an iterative greedy approach based on 
the transition probability of topics from
the content model. It selects $\hat{y}_{T+1} = \arg\max_{s_j \in C_{T+1}}P(\textrm{topic}(s_j)|\textrm{topic}(y_{T}))$ 
at each timestep until the 
length limit is reached. 
This baseline simulates a degenerate version of next-sentence prediction, 
where the choice is based on a single feature at topic level; i.e., the probability of 
transitioning from the topic of the last summary sentence to the topic of the candidate.  Like our model, this baseline has no access to the oracle partial summary, and uses its previous decisions for next sentence selection.

{\bf CHMM-T} is also an iterative greedy approach where the 
evaluation function is the product between topic transition  probability 
(Transition) and topic importance (CHMM).

Apart from the above domain baselines, we also compare with two other types of summaries.

\indent {\bf General} is based on a recent competitive neural network 
based extractive system \cite{cheng-lapata:2016}. This model is designed to be 
domain-general. We trained it on the DailyMail dataset 
\cite{hermann:nips2015}, containing around 200K articles and 
their highlights, without using pretrained embeddings. Our systems are not directly
comparable, because \textsc{NextSum} is trained on much less data, but we show this result to give
an idea of the performance of recent methods.

{\bf Oracle} is the gold-standard extractive summary created from abstracts using 
the mapping method from Section~\ref{sec:oracle-extracts}. It 
represents an upper bound on the performance 
of any extractive summary.



\subsubsection{Results}
Table~\ref{tab:rougeresults} shows the ROUGE-2 F-score results for 
all the systems. The baselines, \textsc{NextSum}$_L$, \emph{oracle} and
\emph{general} 
are fixed length summaries.

\begin{table}
\small{
\begin{tabular}{|l | p{0.8cm} c p{0.8cm} c |}
\hline
{\bf Model}             & \multicolumn{4}{|c|}{\bf ROUGE-2 F-scores}\\
                        & {\sc crime}  & {\sc assassin.} & {\sc bombs} & {\sc mixed}\\ \hline
\multicolumn{5}{|l|}{\textbf{Baselines:}}\\
Lead                    & 0.240        & 0.210           & {\bf 0.250} & 0.232\\ 
CHMM                    & 0.220        & 0.156           & 0.135       & 0.139\\
Transition              & 0.210        & 0.120           & 0.179       & 0.153\\
CHMM-T                  & 0.210        & 0.120           & 0.176       & 0.153\\ \hline
\multicolumn{5}{|l|}{\textbf{Our models:}}\\
\textsc{NextSum}$_{L}$  & {\bf 0.278}  & {\bf 0.227}     & 0.240       & {\bf 0.234}\\
\textsc{NextSum}        & {\bf 0.281}  & {\bf 0.241}     & {\bf 0.250} & {\bf 0.241}\\ \hline
\multicolumn{5}{|l|}{\textbf{Other comparisons:}}\\
General                 & {\bf 0.281}  & 0.201           & 0.237       & 0.225\\
Oracle                  & 0.420        & 0.350           & 0.365       & 0.363\\ \hline
\end{tabular}
\caption{ROUGE-2 F-scores for the generated summaries. The best results for each 
domain are bolded. 
}
\label{tab:rougeresults}
}
\end{table}



  
Among the baselines, we see that the simple lead summary comprising the first $k$ words of the 
source article is the strongest, outperforming domain-trained content model systems in  all the 
domains. The oracle results, however, show that there is still considerable scope for the improvement of automatic systems
performing sentence extraction.  The oracle extractive summary (which was chosen to maximize similarity 
with the abstract) gets close to double the ROUGE score of lead baseline in the {\sc crime} domain. 

Both \textsc{NextSum}$_L$ and \textsc{NextSum} outperform the lead (with statistical significance) 
in all cases except the {\sc bombs} domain. 
Importantly, \textsc{NextSum}, which does automatic 
length prediction, outperforms \textsc{NextSum}$_L$, 
indicating that automatically tailoring summaries to different lengths 
is clearly of value. In 
the next section, we examine this length prediction ability in detail.

Comparing performance across domains, the source articles in {\sc bombs} domain 
are on average longer than the other domains (refer Table \ref{tab:lengthdetails}), which 
could be a reason that content selection performance is lower here. 
This domain also has longer gold standard summaries and the correlation between 
the length of human abstracts and source articles is also the lowest in this domain. 

The \emph{domain-general} system of \newcite{cheng-lapata:2016} is trained on
a much larger general corpus of summary-article pairs. While our results are not directly 
comparable, we see that \textsc{NextSum}'s performance
is competitive with current methods, and since it is based on 
a new outlook and no explicit constraints, it provides much scope for future improvements. 

\subsection{Performance of length prediction}

\textsc{NextSum} requires neither redundancy removal nor length constraints. In this section, 
we show that our system produces summaries of 
varied lengths which correlate with the lengths of human-written summaries
of the same source article. 

\begin{figure}
	\centering
	\includegraphics[width=0.4\textwidth]{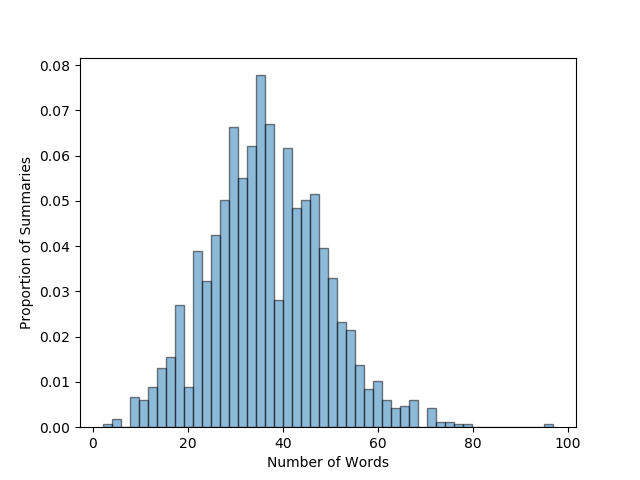}
	\caption{Distribution of lengths (in words) of summaries generated by \textsc{NextSum}.}
	\label{fig:length-distribution}
\end{figure}

Figure~\ref{fig:length-distribution} shows the distribution of the length (in words)
 of \textsc{NextSum} summaries (all domains put together). The generated 
lengths vary greatly, and span the average range covered by the summaries in the 
training data. The majority of lengths are in the 30 to 50 words limit. 
Hence  \textsc{NextSum} is specializing summary lengths to cover a wide range. 


Next, we measure how well these summary lengths correlate with the lengths of the \emph{human-written abstracts}. 
Table~\ref{tab:lengthcorrelations} shows
 the Kendall Tau correlation (corrected for ties) between length in words of the \textsc{NextSum} summary 
and the length of the abstract for the same source. 


\textsc{NextSum}'s summary lengths correlate fairly well with those of the abstracts, leading to significant
numbers in all the domains and the mixed case. Again, the length prediction is worse on 
the {\sc Bombs} domain compared to the rest.
Overall,
this result shows promise that we can develop
summarization systems which automatically tailor
their content based on properties of the source.

\begin{table}
  \centering{
    \small{
	\begin{tabular}{|l r|}
          \hline
          {\bf Domain}    & {\bf Tau} \\ \hline
          {\sc crime}     & 0.46  \\
          {\sc assassin.} & 0.40  \\
          {\sc bombs}     & 0.28 \\
          {\sc mixed}     & 0.32 \\ \hline
        \end{tabular}
        \caption{Kendall Tau correlation between length (in words) of \textsc{NextSum} summaries and 
human abstracts.}
\label{tab:lengthcorrelations}
}}
\end{table}

\section{Conclusion}
In this work, we have presented the first summarization system which 
integrates content selection, summary length prediction, and redundancy removal. Central to this system is the use of 
a next-sentence prediction system which learns summary-internal discourse transitions.
We show that \textsc{NextSum} outperforms a number of baselines on ROUGE-2 F-scores even when the 
summary length is not provided to the system. Furthermore, the lengths of the predicted 
summaries correlate positively with the lengths of human-written abstracts, indicating that our 
method implicitly captures some aspect of how much summary-worthy content is present 
in the source article.

In future work, we plan to investigate whether this approach also leads to more coherent summaries.
This issue will be especially important in the multi-document setting, which we would also like to investigate using an extension of our model.



\bibliography{ssum}
\bibliographystyle{acl_natbib}

\end{document}